\documentclass[letterpaper, 10 pt, conference]{ieeeconf}  % Comment this line out if you need a4paper
\pdfoutput=1
\IEEEoverridecommandlockouts                              % This command is only needed if 
\usepackage{gensymb}
\usepackage{cclicenses, graphicx}
\usepackage{amsmath}
\usepackage{array}
\usepackage[keeplastbox]{flushend}

\newcolumntype{C}[1]{>{\centering\arraybackslash}m{#1}}

\title{\LARGE \bf Intuitive Hand Teleoperation by Novice Operators\\ Using a Continuous Teleoperation Subspace}

\author{Cassie Meeker$^{1}$, Thomas Rasmussen$^{1}$ and Matei Ciocarlie$^{1}$%
\thanks{*
This work was supported in part by the 
ONR Young Investigator Program award N00014-16-1-2026.}%
\thanks{$^{1}$Department of Mechanical Engineering, Columbia University, New York, NY 10027, USA.}%
\thanks{\hspace{-3mm}{\tt\small \{cgm2144, matei.ciocarlie\}@columbia.edu}}%
\thanks{\hspace{-3mm}{\tt\small \{tomras12\}@gmail.com}}
}

\begin{document}

\maketitle
\thispagestyle{empty}
\pagestyle{empty}

%%%%%%%%%%%%%%%%%%%%%%%%%%%%%%%%%%%%%%%%%%%%%%%%%%%%%%%%%%%%%%%%%%%%%%%%%%%%%%%%
\begin{abstract}
Human-in-the-loop manipulation is useful in when autonomous grasping is not 
able to deal sufficiently well with corner cases or cannot operate fast 
enough. Using the teleoperator's hand as an input device can provide an 
intuitive control method but requires mapping between pose spaces which may 
not be similar. We propose a low-dimensional and continuous teleoperation 
subspace which can be used as an intermediary for mapping between different 
hand pose spaces. We present an algorithm to project between pose space and 
teleoperation subspace. We use a non-anthropomorphic robot to 
experimentally prove that it is possible for teleoperation subspaces to 
effectively and intuitively enable teleoperation. In experiments, novice 
users completed pick and place tasks significantly faster using 
teleoperation subspace mapping than they did using state of the art 
teleoperation methods. 

\end{abstract}
%%%%%%%%%%%%%%%%%%%%%%%%%%%%%%%%%%%%%%%%%%%%%%%%%%%%%%%%%%%%%%%%%%%%%%%%%%%%%%%%

\section{Introduction}

In unstructured environments that require unscripted and complex
manipulation, it is often useful to utilize human-in-the-loop
manipulation in lieu of autonomous grasping. When a large number
of possible scenarios and objects can be encountered, human
cognition makes decisions faster and deals with corner cases
better than autonomous algorithms. 

Teleoperation which harvests the movement of the teleoperator's 
hand to control a robot hand can provide an intuitive and user-friendly 
interface~\cite{ferre2007}. This type of teleoperation 
uses hand motions which are already familiar to the user, instead 
of requiring knowledge of external control hardware. However, 
teleoperation based on the operator's hand movement requires mapping 
between the pose spaces of the two hands of interest. 
Intuitive teleoperation mappings are desirable because they 
help teleoperators, particularly novice teleoperators, to complete tasks 
in a safe and timely manner.

Robot hand designs which are fully-actuated and anthropomorphic allow 
for an intuitive mapping between hands and thus are
intuitive for a human to teleoperate; however, the hardware tends to be fragile
and expensive. In contrast, non-anthropomorphic hands have proven to be 
robust and versatile in unstructured environments. However, finding an easy
or intuitive mapping between the human hand and a non-anthropomorphic 
robot hand can be difficult, due to the different joints,
different axes, different numbers of fingers, or any number of
dissimilarities between the hands.

In this paper, we seek to create an intuitive mapping between the human hand
and a fully actuated non-anthropomorphic robot hand that enables
effective real-time teleoperation for novice users. 

The method we propose uses a subspace relevant to teleoperation
as an intermediary between the pose spaces of two different hands. 
Our method enables teleoperation by projecting the pose of the master hand 
into the defined teleoperation subspace, which it shares with the slave 
hand, and then projecting from the teleoperation subspace into the pose
space of the slave hand. 

Unlike traditional pose mapping, we avoid unexpected movements by using a 
continuous subspace as a basis for mapping instead of interpolation between 
discrete poses. Our mapping is independent of the master-slave pairing, so 
the mapping between teleoperation subspace and pose space of a robot does 
not have to be redefined with every new human teleoperator. Furthermore, 
the teleoperation subspace is low dimensional, which allows for the future 
possibility of simple control mechanisms, such as cursor control or 
electromyography (EMG) based controls, although these are not explored in 
this work. 

Our main contributions are as follows: we introduce a continuous, low-dimensional 
teleoperation subspace as an intuitive way to map human to 
robot hand poses for teleoperation. We posit that this method allows for 
intuitive teleoperation as long as both the master and the slave hand poses 
can be projected into this subspace. We provide an empirical method for 
achieving this projection, and experimentally prove that it is effective 
and intuitive using a robot hand with highly non-anthropomorphic 
kinematics. Our method allows novice teleoperators to pick and place 
objects significantly faster than state of the art teleoperation mapping 
methods.

\section{Related Work}

Conventional teleoperation methods are divided into three main categories: 
joint mapping, fingertip mapping, and pose mapping. 

Joint mapping (also called joint-to-joint mapping) is used when the slave 
hand has similar kinematics to the human controller~\cite{liarokapis2013}. 
If the human and robot joints have a clear correspondence, the human joint 
angles can be imposed directly onto the robot joints with little or no 
transformation~\cite{cerulo2017}. This mapping is most useful for power 
grasps~\cite{chattaraj2014}, and is limited if the robot hand is non-anthropomorphic.

Fingertip mapping (also called point-to-point mapping) is the most common 
teleoperation mapping method. Forward kinematics transform human joint 
angles into Cartesian fingertip positions. These undergo scaling to find 
the desired robot Cartesian fingertip positions and then inverse kinematics 
determine robot joint angles. This mapping is useful for precision grasps~\cite{chattaraj2014}. 
When the robot has less than five fingers, the extra 
human fingers are ignored~\cite{peer2008}. Error compensation can find the 
closest fit when human and robotic workspaces are incompatible~\cite{rohling1993}.
However, teleoperation using this method is difficult when 
the workspaces of the human and robotic fingers are not similar. 

An alternative to fingertip mapping is virtual object mapping, which uses 
the relative distances between fingertips. Fingertip mapping uses the 
distance between the master fingertips and inverse kinematics to calculate 
joint angles that place the slave fingertips at the same relative distance. 
This method can be used in both 2-D~\cite{griffin2000} and 3-D~\cite{wang2005}\cite{gioioso2013} 
grasping scenarios. The relative distances 
between fingertips are often calculated based on Cartesian fingertip 
positions, so virtual object mapping is similar to fingertip mapping. 
Virtual object mapping is useful for tasks which involve dexterous 
manipulation~\cite{salvietti2013}; however it is often unsuitable for tasks 
which require different grasp types or which involve irregularly shaped 
objects. This mapping was unsuitable for our experiments because of the 
variety of objects we wished to pick and place. Virtual object mapping can 
generalize to virtual objects of any shape~\cite{salvietti2017}, assuming 
that contact points for the slave and master hands can be tracked. This is 
difficult without haptic devices, which we did not use in this work.

Pose mapping attempts to replicate the pose of the human hand with a robot 
hand, which is appealing because, unlike fingertip and joint mapping, it 
attempts to interpret the function of the human grasp rather than replicate 
hand position. Pao and Speeter define transformation matrices relating 
human and robot poses using least squared error compensation when this 
transformation is not exact~\cite{pao1989}. Others use neural networks to 
identify the human pose and map the pose to a robot either through another 
neural network~\cite{ekvall2004} or pre-programmed joint-to-joint mapping~\cite{wojtara2004}. 
Outside of a discrete set of known poses, pose mapping 
can lead to unpredictable hand motions and is usually used when only simple 
grasping poses are required. Furthermore, the above mappings that use 
neural networks require classification of the human hand pose before it is 
mapped to the robotic hand. If this classifier misidentifies the human 
pose, the robot hand will move in undesirable ways. Our method also 
attempts to replicate hand shape, rather than fingertip or joint positions, 
making it most similar to pose mapping, but we do not require discrete 
classification of human pose before mapping.

This paper introduces a low-dimensional mapping. Other methods that define 
grasping in a low dimensional space include postural synergies, which are 
low dimensional and continuous~\cite{santello1998}. Just as synergies move 
the description of human hand position from discrete, static poses~\cite{feix2009} 
into a continuous space, we seek to allow pose mapping between 
the human and robotic hand to be continuous instead of interpolating 
between discrete poses. 

In the field of autonomous grasp planning, finding synergies for robotic 
hands based on human synergies can inform grasp 
planning~\cite{ficuciello2012}\cite{palli2014}\cite{geng2011}. Other works show 
synergies to also be an effective control for teleoperation. Jenkins 
demonstrated a low dimensional control which could potentially be used to 
teleoperate robotic systems with cursor control~\cite{jenkins2008}. To our 
knowledge, this has only been tested in simulation. Brygo, et al. 
translated postural synergies from joint space to fingertip Cartesian space 
to control teleoperation~\cite{brygo2017}. This work only considers the 
first postural synergy, and is most appropriate for underactuated hands. 
Kim, et al. demonstrated a synergy level controller which uses multiple 
postural synergies to enable teleoperation~\cite{kim2016}. This method 
calculates the synergies of the robot hand through pose mapping, which 
could have discontinuities. They calculated robotic synergy coefficients 
based, in part, on the rate of change for each synergy coefficient. Our 
method does not have discontinuities and does not have a temporal component.

Other works use low dimensional latent variables which are not based on 
synergies to approximate human poses in non-anthropomorphic models. These 
latent variables have enabled both the animation of non-anthropomorphic 
creatures~\cite{yamane2010} and teleoperation. Gaussian process latent 
variable models (GP-LVM) can enable teleoperation of humanoid robots. In 
some formulations, the latent space changes with every different master-slave 
pairing~\cite{shon2006}. In other formulations, multiple robots and a 
human share the same latent space~\cite{delhaisse2017}. These latent 
variable models require the user to generate observations where the master 
and slave poses are correlated in order to train the GP-LVM model. A given 
GP-LVM model requires the user to generate tens, and even up to hundreds of 
correlated poses for training. Although we are inspired by a similar desire 
to find shared subspaces between robotic and human hands, our method only 
requires the user to generate one correlated pose. We also require two 
other vectors for each hand, but we outline a simple way to calculate those 
vectors that does not require consideration of the master-slave pairing.

\section{Teleoperation Subspace} \label{method}

\begin{figure}[t]
\centering
\begin{tabular}{r}
\includegraphics[trim=5.37cm 7.7cm 2.1cm 4.5cm, clip, width=1\linewidth]{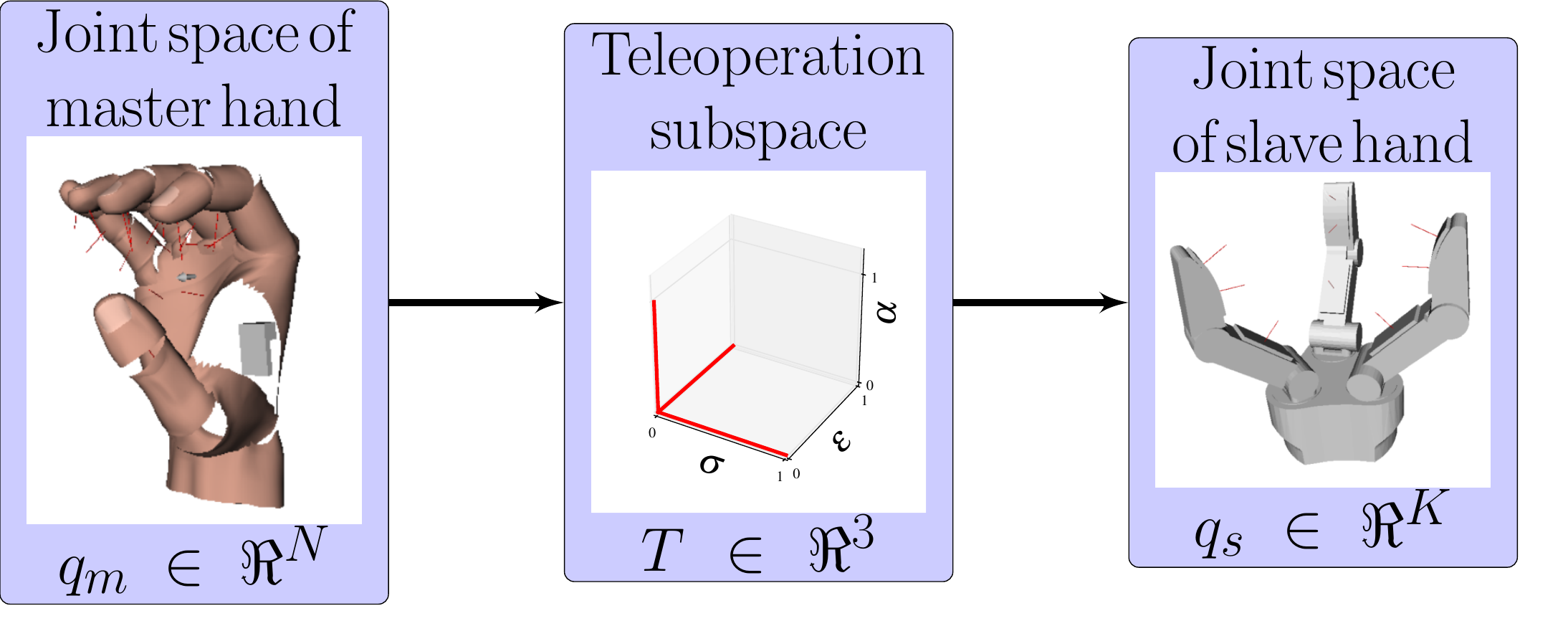}
\end{tabular}
\caption{Steps to enable real time teleoperation using teleoperation subspace}
\label{flow_chart}
\vspace{-4mm}
\end{figure}

As a general concept, we posit that, for many hands, a three dimensional 
space $T$ isomorphic to $\Re^3$ can encapsulate the range of movement 
needed for teleoperation. The three basis vectors of $T$ have an intuitive 
correspondence with hand shape. One corresponds with how far apart the 
fingers are spread, another with the size of the object a hand can grasp, 
and another with how curled the fingers are. We refer to these basis 
vectors as $\boldsymbol\alpha, \boldsymbol\sigma$, and $\boldsymbol\epsilon$, 
respectively.

We chose these bases using on intuition, guided by Santello's research of 
postural synergies~\cite{santello1998}. Since Santello, et al. used principle 
component analysis (PCA), a linear dimension reduction method, to find 
postural synergies, we also assume that mapping between pose space and 
teleoperation subspace is linear. 

We assume that many hands will be able to
project their pose spaces into $T$. If this projection is possible, $T$ is 
embedded as a subspace in the pose space of the hand. $T$ is thus a subspace 
``shared" by all hands that can project their pose space into $T$. If the user 
can construct a projection matrix which projects pose space to teleoperation 
subspace in a meaningful way, our method will enable teleoperation. 
Experimentally, we prove that $T$ is relevant for teleoperation for at least 
the human hand and a non-anthropomorphic robot hand, similar to the Schunk 
SDH. We theorize that $T$ is also relevant to teleoperation for other hands.

To teleoperate using $T$, there are two steps:
\begin{enumerate}
\item Given joint values of the master hand $h_m$, find the equivalent pose $t$
in teleoperation subspace $T$
\item Given $t$ computed above, find the joint values 
of the slave hand $h_s$, and move $h_s$ to these values
\end{enumerate}

In order to enact the teleoperation steps, we must first project
between $T$ and the relevant pose spaces.

\subsection{Projecting between Pose Space and Teleoperation Subspace}\label{changing_spaces}

We define an empirical projection method from pose space to teleoperation 
subspace. For a given hand with $N$ joints, projecting from joint space $q \in \Re^N$ 
(here we use pose space and joint space interchangeably) into 
teleoperation subspace $T$ requires an origin pose $o \in \Re^{N}$, a 
projection matrix $A \in \Re^{N \times 3}$, and a scaling factor $\delta \in \Re^{3}$.
%%%%%%ORIGIN
\subsubsection{Origin $o$}\label{origin}
To project between joint space and $T$, we require a hand-specific, ``neutral" origin 
pose $o \in \Re^N$.
\begin{equation}
o = [o_1, o_2, ... , o_N]
\end{equation}

\begin{figure}[t]
\centering
\begin{tabular}{cccc}
\includegraphics[trim=7cm 2cm 7cm 8cm,clip, width=.22\linewidth]{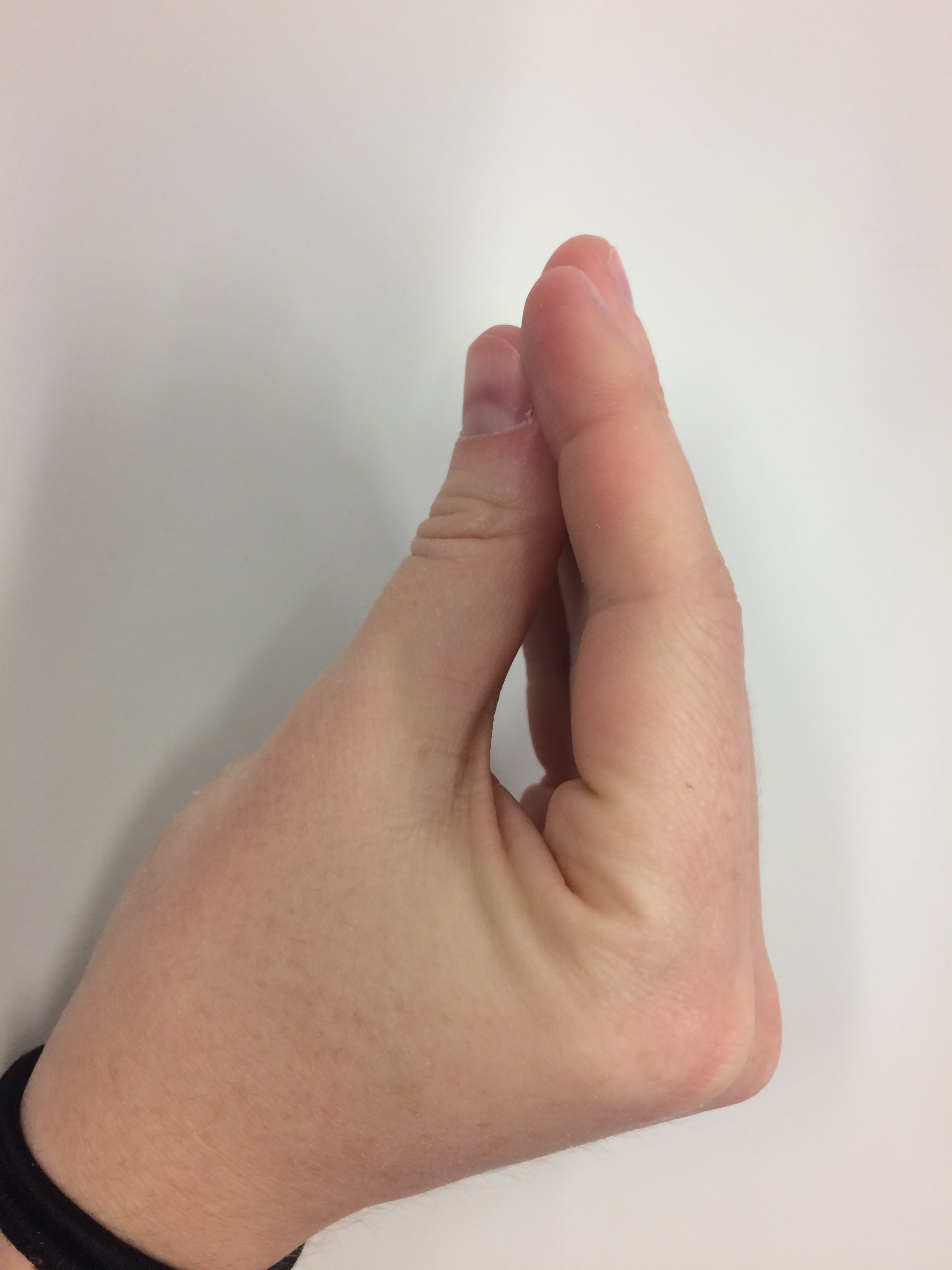}\hspace{.1cm}
\includegraphics[trim=0cm 0cm 0cm 0cm,clip, width=.22\linewidth]{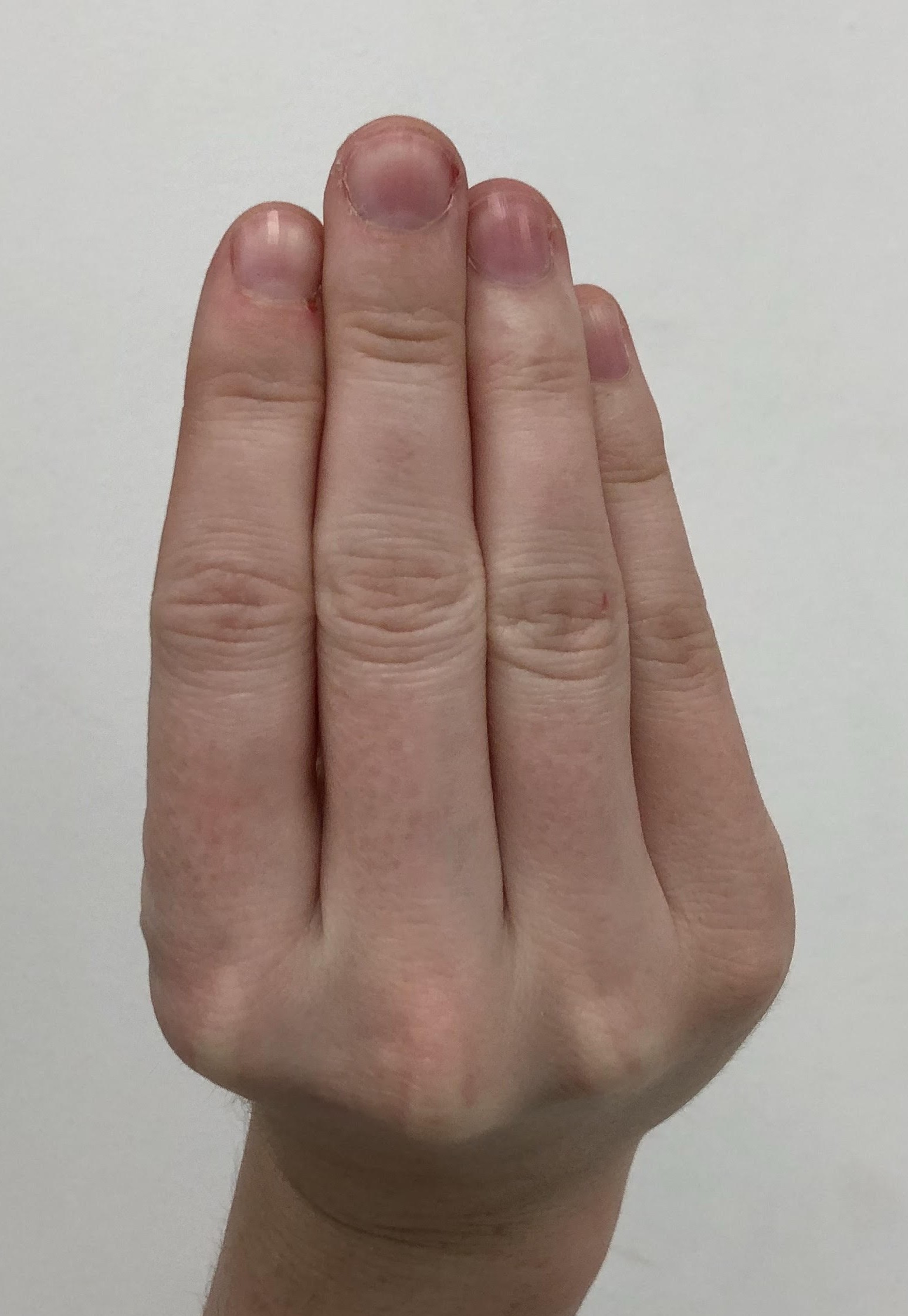}\hspace{.1cm}
\includegraphics[trim=32cm 2cm 26.5cm 2cm,clip, width=.22\linewidth]{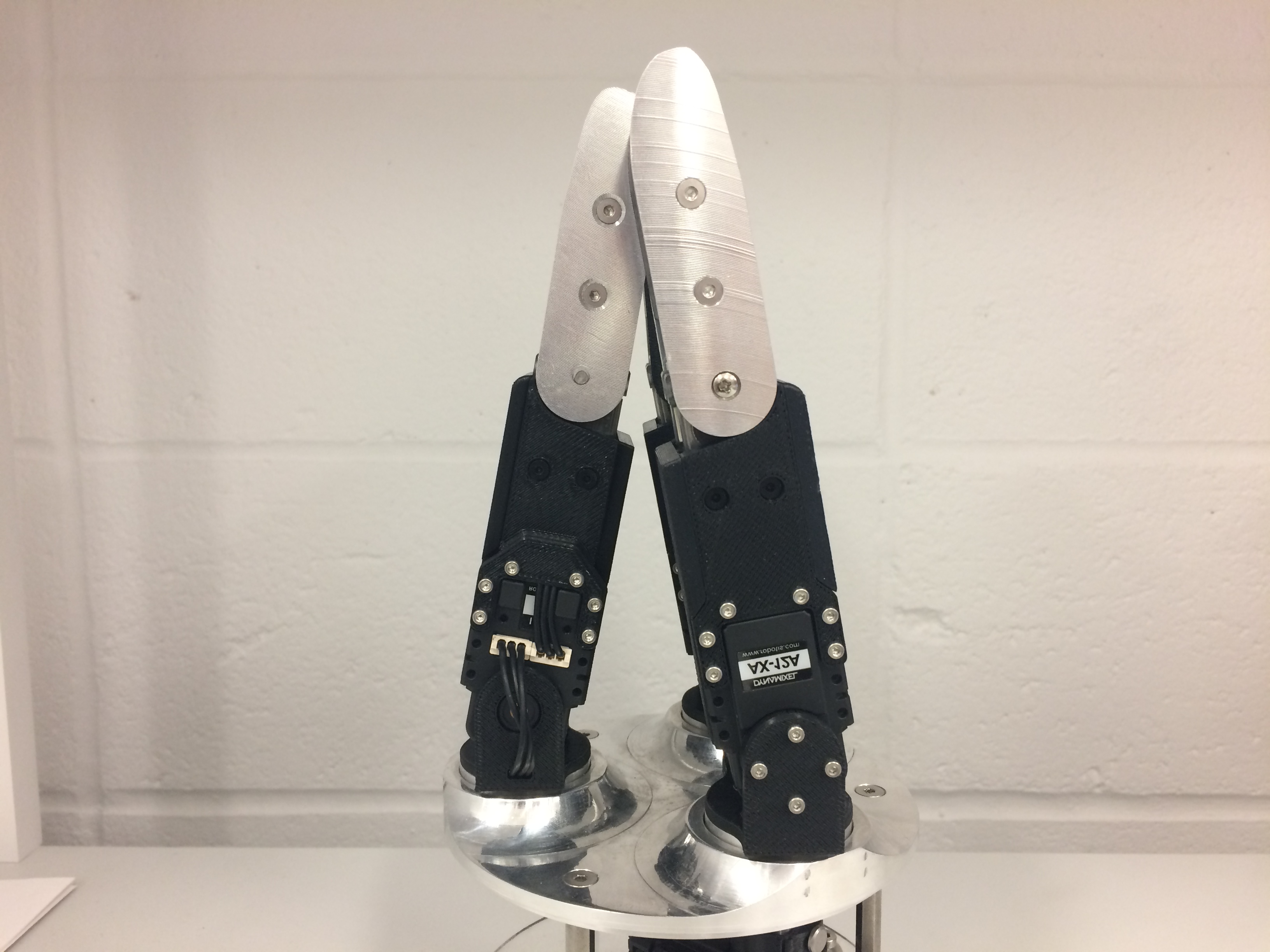}\hspace{.1cm}
\includegraphics[trim=28cm 0cm 32cm 6cm,clip, width=.22\linewidth]{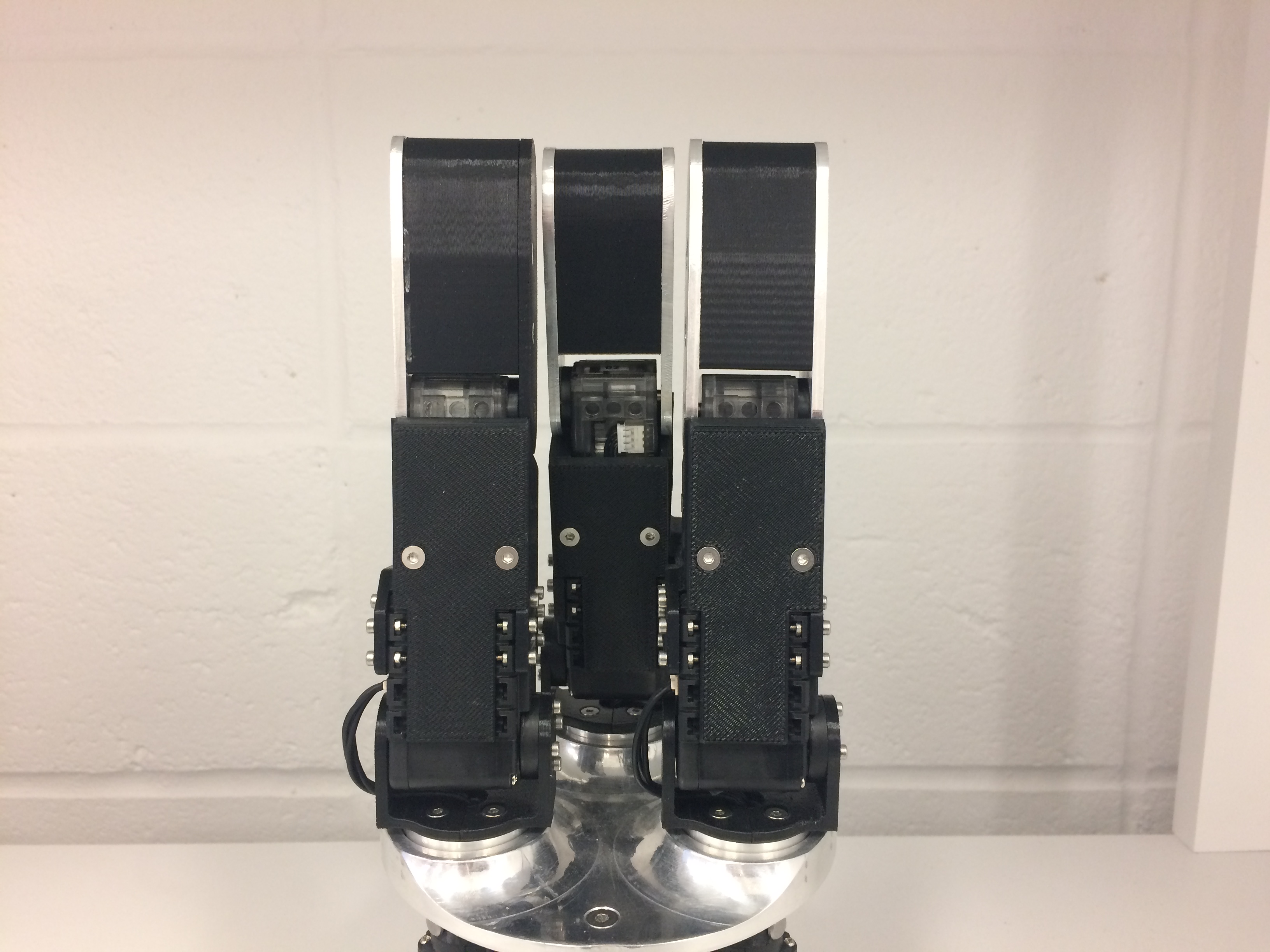}\\
\parbox[c]{.44\linewidth}{\footnotesize \centering Origin pose of the human hand.}\hspace{.1cm}
\parbox[c]{.44\linewidth}{\footnotesize \centering Origin pose of the robotic hand.}\hspace{.1cm}
\vspace{-2mm}
\end{tabular}
\caption{Origin poses of two example hands.}
\label{origin_pose}
\vspace{-4mm}
\end{figure}

This represents a hand position which will standardize the data as
we project between joint space and $T$. The origin pose of 
the master is arbitrary; however, it is crucial that the origin 
pose of the slave corresponds to the master's origin 
(i.e. the two hands should assume approximately the same shape 
while positioned at their respective origins). 
Figure~\ref{origin_pose} shows the pose we chose for the human 
hand and the custom-built robotic hand in our experiments. 

%%ORIGIN POSES FIGURE

%%Projection Matrix
\subsubsection{Projection Matrix $A$}\label{projection_matrix}
The projection matrix $A \in \Re^{N \times 3}$ is hand
specific and consists of three basis vectors $\psi \in \Re^N$. Each
$\psi$ is a projection of one of $T$'s basis vectors in pose space. 
The $\psi$s which correspond to $\boldsymbol\alpha, \boldsymbol\sigma$, 
and $\boldsymbol\epsilon$ are referred to as $\boldsymbol\psi_\alpha$, 
$\boldsymbol\psi_\sigma$, and $\boldsymbol\psi_\epsilon$, respectively. 
\begin{eqnarray}
A = [\psi_\alpha, \psi_\sigma, \psi_\epsilon] \\ 
\psi_\alpha = [\psi_\alpha{_1}, \psi_\alpha{_2}, ... , \psi_\alpha{_N}]^\top \\
\psi_\sigma = [\psi_\sigma{_1}, \psi_\sigma{_2}, ... , \psi_\sigma{_N}]^\top \\
\psi_\epsilon = [\psi_\epsilon{_1}, \psi_\epsilon{_2}, ... , \psi_\epsilon{_N}]^\top
\end{eqnarray}

For each joint, the user determines the basis vector in $T$ to which a 
joint has the most intuitive correspondence, and then sets that element 
equal to 1 in the $\psi$ which is a projection of that basis vector. The 
user sets joints which adduct the fingers to 1 in $\psi_\alpha$, joints 
which open the hand to 1 in $\psi_\sigma$, and joints which curl the fingers 
to 1 in $\psi_\epsilon$. This is a winner take all approach, so a joint may 
only contribute to a single $\psi$. This is most relevant for underactuated 
hands where a single joint could simultaneously serve to open the hand and 
spread the fingers, but can only be non-zero in one $\psi$. After the $\psi$s 
are built, the user normalizes each $\psi$ to create $A$.

As an illustrative example, we build $A$ for the robotic slave hand used in 
our experiments. The robotic hand has eight joints - the thumb ($f0$) has a 
proximal and distal joint, and the two opposing fingers ($f1$ and $f2$) have 
proximal, distal and adduction joints. We define the vector of joints as: 
\begin{equation}
j = [f0_{prox}, f0_{dis}, f1_{ad}, f1_{prox}, f1_{dis} , f2_{ad}, f2_{prox}, f2_{dis}]
\end{equation}

For this hand, joints which spread the fingers are the adductors at $j[2]$ and 
$j[5]$. These joints have the most intuitive correspondence to $\alpha$ so they 
are set to 1 in $\psi_\alpha$:
\begin{equation}
\psi_\alpha = [0, 0, 1, 0, 0, 1, 0, 0]
\end{equation}
Changing the values at the proximal joints allows the hand to grasp objects of 
varying sizes, meaning they correspond to $\sigma$. Since $j[0]$, $j[3]$ and 
$j[6]$ represent the proximal joints, we set the values at these indices as 1 in 
$\psi_\sigma$:
\begin{equation}
\psi_\sigma = [1, 0, 0, 1, 0, 0, 1, 0]
\end{equation}
For the robotic hand, changing the values at the distal joints will curl the 
fingers. The distal joints correspond to $j[1]$, $j[4]$ and $j[7]$, so we set 
the values at these indices as 1 in $\psi_\epsilon$:
\begin{equation}
\psi_\epsilon = [0, 1, 0, 0, 1, 0, 0, 1]
\end{equation}
Finally, we normalize the above $\psi$s in order to create $A$.

Even though our method requires user intuition to build the projection matrix $A$, 
the process of building $A$ as described above is fairly simple. Experimentally, 
we prove that these basic calculations are sufficient to meaningfully project 
pose space into $T$ in a way that enables teleoperation.

\subsubsection{Scaling Factor $\delta$}\label{scaling_factor}

We wish to normalize such that any configuration in pose space will project to 
a pose in $T$ whose value is less than or equal to 1 along each of the basis 
vectors. We therefore require a scaling factor $\delta \in \Re^{3}$ to normalize 
the projection:
\begin{equation}
\delta = [\delta_\alpha, \delta_\sigma, \delta_\epsilon]
\end{equation}

\begin{figure}[t]
\centering
\setlength{\tabcolsep}{-1mm}
\begin{tabular}{cccc}
\includegraphics[trim=13cm 2cm 7cm 0cm,clip, width=.22\linewidth]{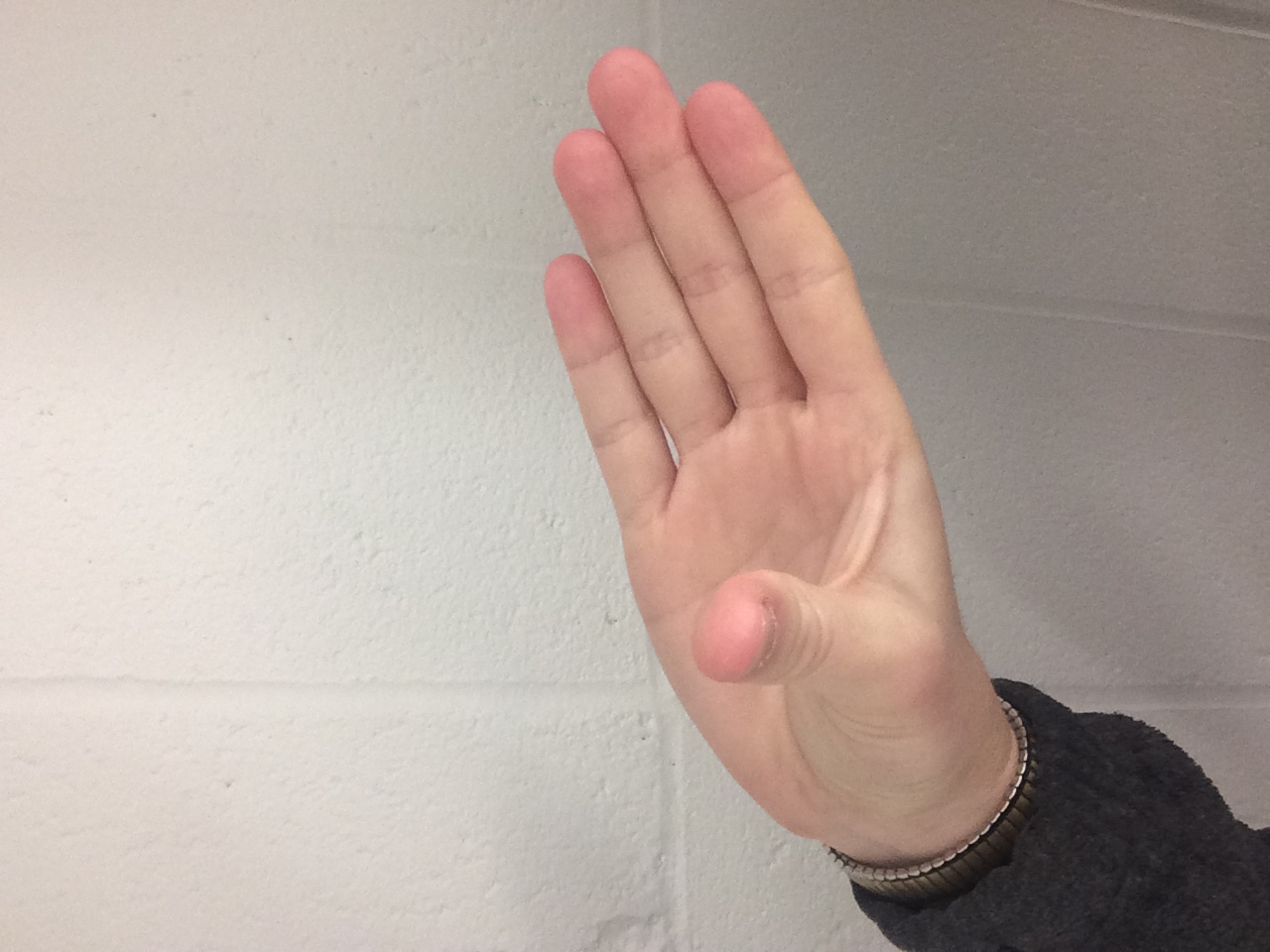} \hspace{1mm}
\includegraphics[trim=13cm 2cm 7cm 0cm,clip, width=.22\linewidth]{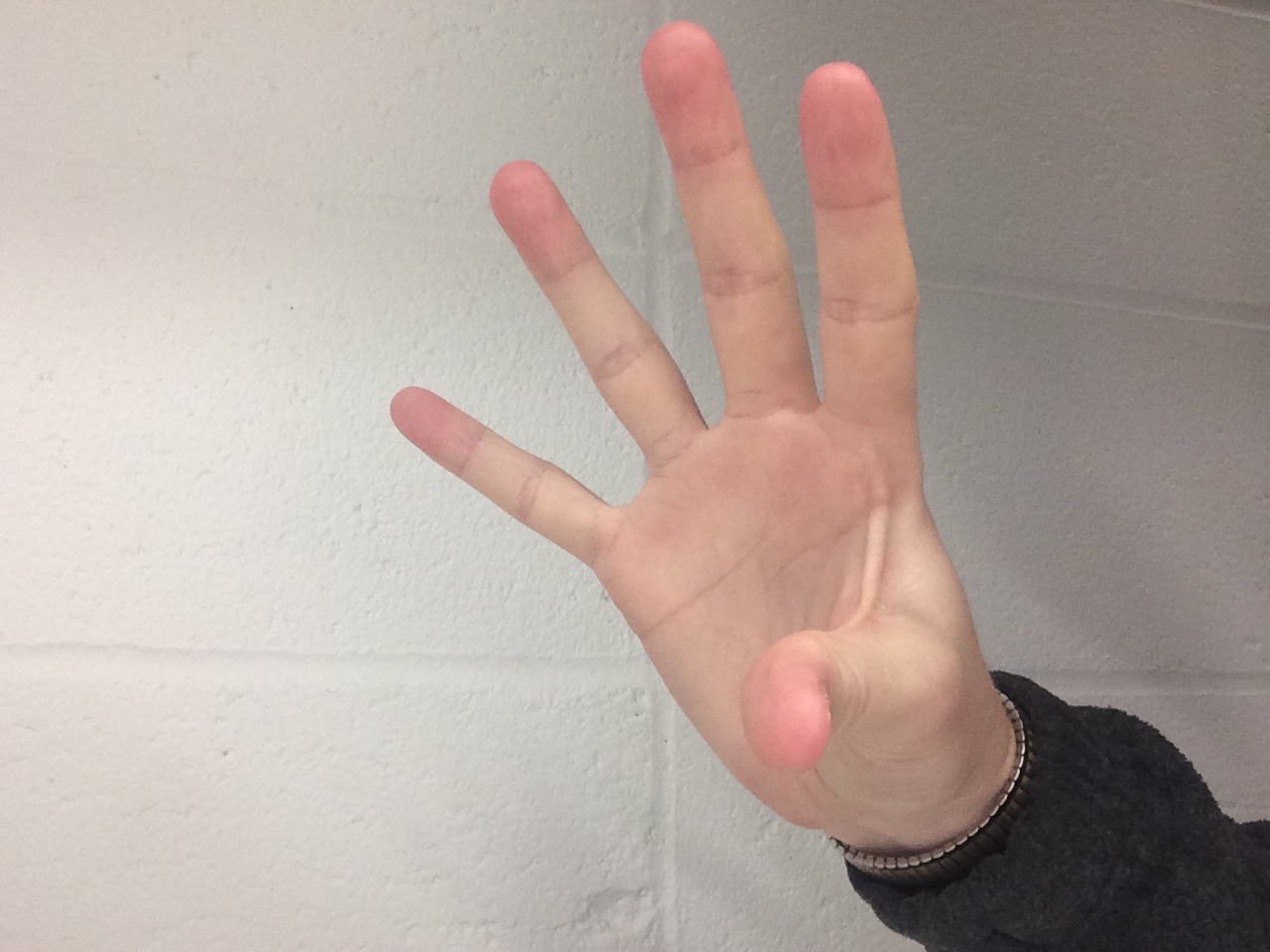} \hspace{1mm}
\includegraphics[trim=16cm 1.9cm 7cm 4cm,clip, width=.22\linewidth]{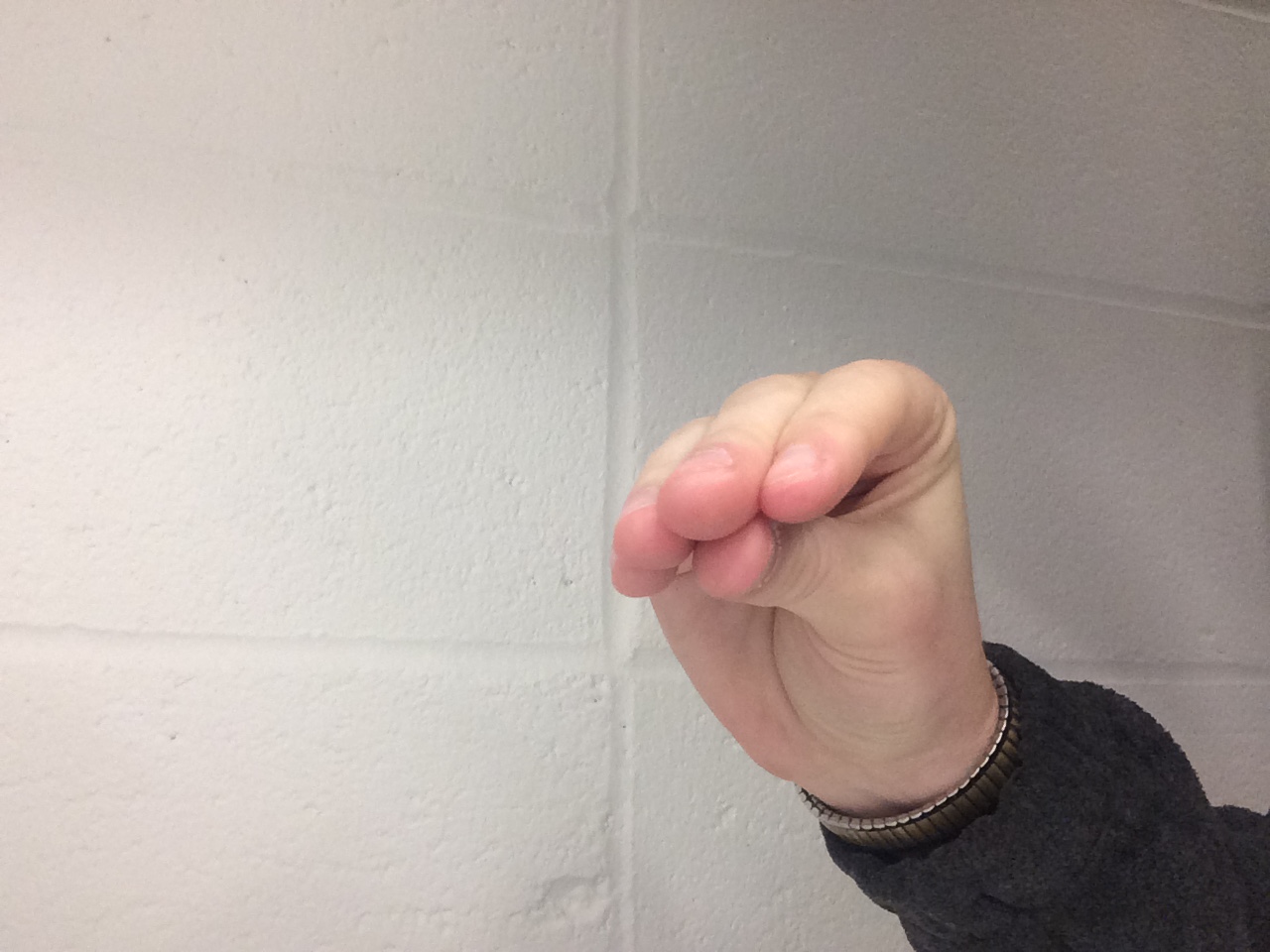} \hspace{1mm}
\includegraphics[trim=13cm 1cm 7cm 1cm,clip, width=.22\linewidth]{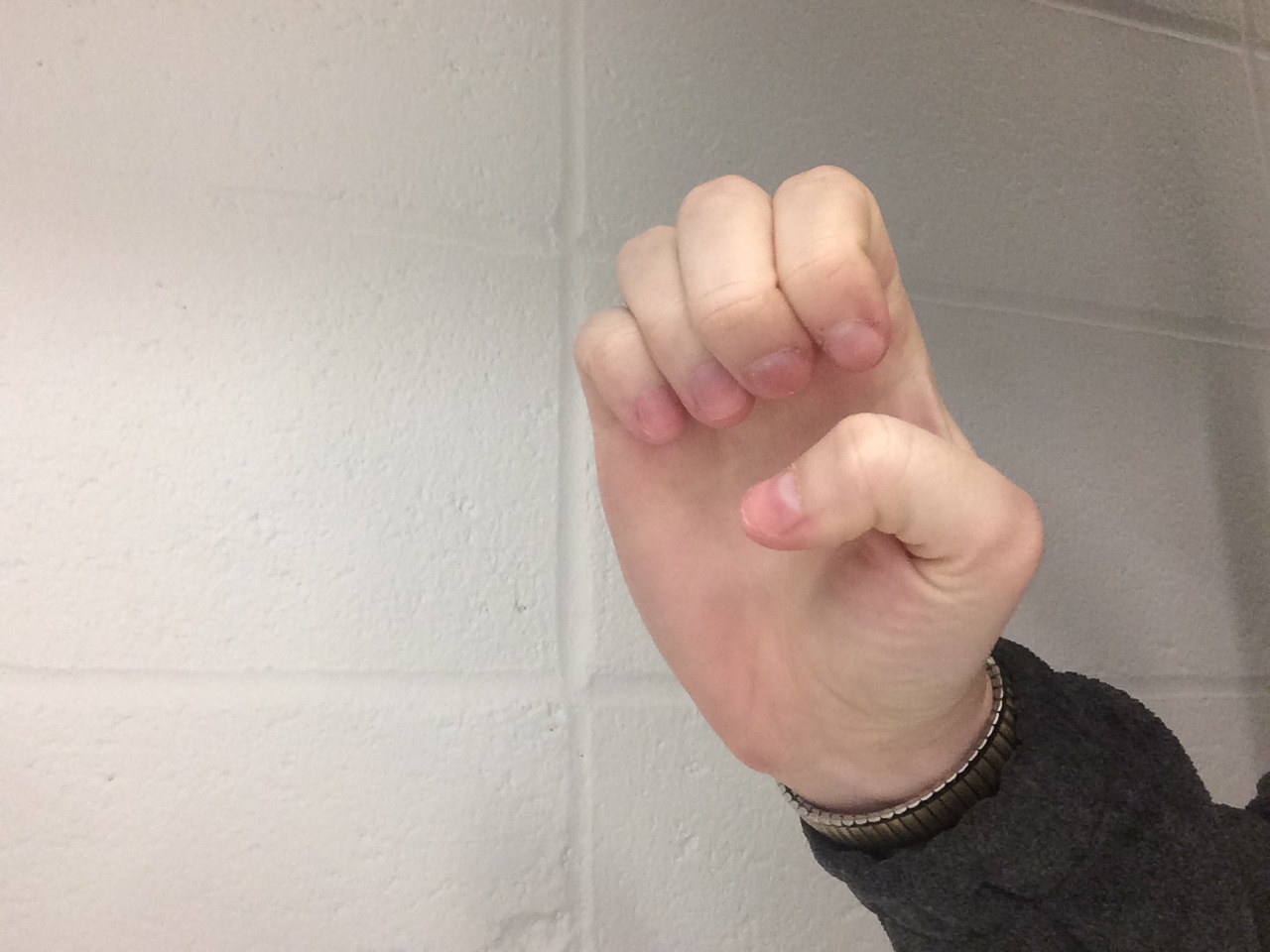}\\
\parbox[c]{.22\linewidth}{\footnotesize \centering Pose for maximum along $\boldsymbol\sigma$, minimum along $\boldsymbol\alpha$.}  \hspace{1mm}
\parbox[c]{.22\linewidth}{\footnotesize \centering Pose for maximum along $\boldsymbol\alpha$.}  \hspace{1mm}
\parbox{.22\linewidth}{\footnotesize \centering Pose for minimum along $\boldsymbol\epsilon$, and minimum along $\boldsymbol\sigma$.} \hspace{1mm}
\parbox{.22\linewidth}{\footnotesize \centering Pose for maximum along $\boldsymbol\epsilon$.} \\
\end{tabular}
\caption{Poses required from the human user to demonstrate maximum and minimum of the unscaled teleoperation subspace projection.}
\label{calibration_poses}
\vspace{-4mm}
\end{figure}

To calculate $\delta$, we evaluate poses which illustrate the extrema of the 
hand's kinematic limits along the basis vectors. For example, the maximum and 
minimum values along $\boldsymbol\sigma$ are illustrated by projecting poses 
where the hand is holding the largest object possible and the smallest object 
possible from pose space into $T$. It is up to the user to determine poses which 
illustrate the full range of values for each basis vector. Figure~\ref{calibration_poses} 
shows the poses which demonstrate these ranges for the human hand.

Once we select the illustrative poses for the hand, we project these poses 
from pose space into $T$ using $t = (q-o) \cdot A$, where $t \in T$. From 
this set of poses in $T$, we find the minimum and maximum values along each 
axis. Along $\boldsymbol\alpha$, the minimum and maximum are referred to as 
$\alpha_{min}$ and $\alpha_{max}$, respectively. From these values, we calculate 
$\delta_{\boldsymbol\alpha}$ as:
\begin{eqnarray}
\alpha_{range} = abs(\alpha_{max}) + abs(\alpha_{min}) \\
\delta_{\boldsymbol\alpha} = 
	  \begin{cases}
         0 & \text{if $\alpha_{range} = 0$ } \\
         1/\alpha_{range} & \text{otherwise}
      \end{cases}
\end{eqnarray}
Finding $\delta_{\boldsymbol\sigma}$ and $\delta_{\boldsymbol\epsilon}$ uses 
the same calculation.

%%CALIBRATION POSES FIGURE

$\delta$ normalizes the projection from pose space to $T$; however, to project 
from $T$ back to pose space, we require an inverse scaling factor $\delta^*$:
\begin{eqnarray}
   \delta^* = [\delta^*_\alpha, \delta^*_\sigma, \delta^*_\epsilon] \\
   \delta^*_\alpha = 
      \begin{cases}
         0 & \text{if $\delta_\alpha = 0$ } \\
         1/\delta_\alpha & \text{otherwise}
      \end{cases}
\end{eqnarray}
where we find $\delta^*_\sigma$ and $\delta^*_\epsilon$ with similar calculations.

\subsubsection{A Complete Projection Algorithm}
To project between teleoperation subspace $T$
and joint space $q$, we use the hand-specific matrix $A$, the
origin $o$, and the scaling factor $\delta$:
%\vspace{-2mm}
\begin{eqnarray}
t = ((q-o) \cdot A) \odot \delta  \label{eq_to_subspace}\\
q = ((t \odot \delta^*) \cdot A^\top) + o \label{eq_from_subspace}
\end{eqnarray}
where, $\odot$ represents element-wise multiplication.

To use $T$ for teleoperation, Eq.~(\ref{eq_to_subspace}) projects 
the master hand's pose space into the shared
teleoperation subspace and then Eq.~(\ref{eq_from_subspace})
projects from the shared teleoperation subspace
into the slave hand's pose space.

So, given the joint angles of the master hand, we are able to
calculate the joint angles of the slave hand using:
\begin{eqnarray}
q_s = (((q_m-o_m) \cdot A_m) \odot \delta_m \odot \delta^*_s) \cdot A_s^\top + o_s
\end{eqnarray}

\section{Experiments}

To show that the proposed method is intuitive for novice users,
experiments were performed with five healthy subjects. Two subjects 
were female, three subjects were male, all were
aged between 23 and 28, and all were novice robot
teleoperators. Subjects gave their informed consent 
and the study was approved by the Columbia University IRB.

We compared our teleoperation method with two state of the art teleoperation
techniques - joint mapping and fingertip mapping. These methods were chosen 
from the state of the art as being the most applicable to our problem: intuitive 
for novice users, and able to use different grasping types (power and precision, 
for example) to grasp objects of various shapes and sizes. We used time to complete 
a pick and place task as a metric for usability and intuitiveness.

\subsection{Experimental Setup}
%%SETUP FIGURE
\begin{figure}[t]
\centering
\begin{tabular}{r}
\includegraphics[trim=0cm 0cm 30cm 0cm, clip,width=.58\linewidth]{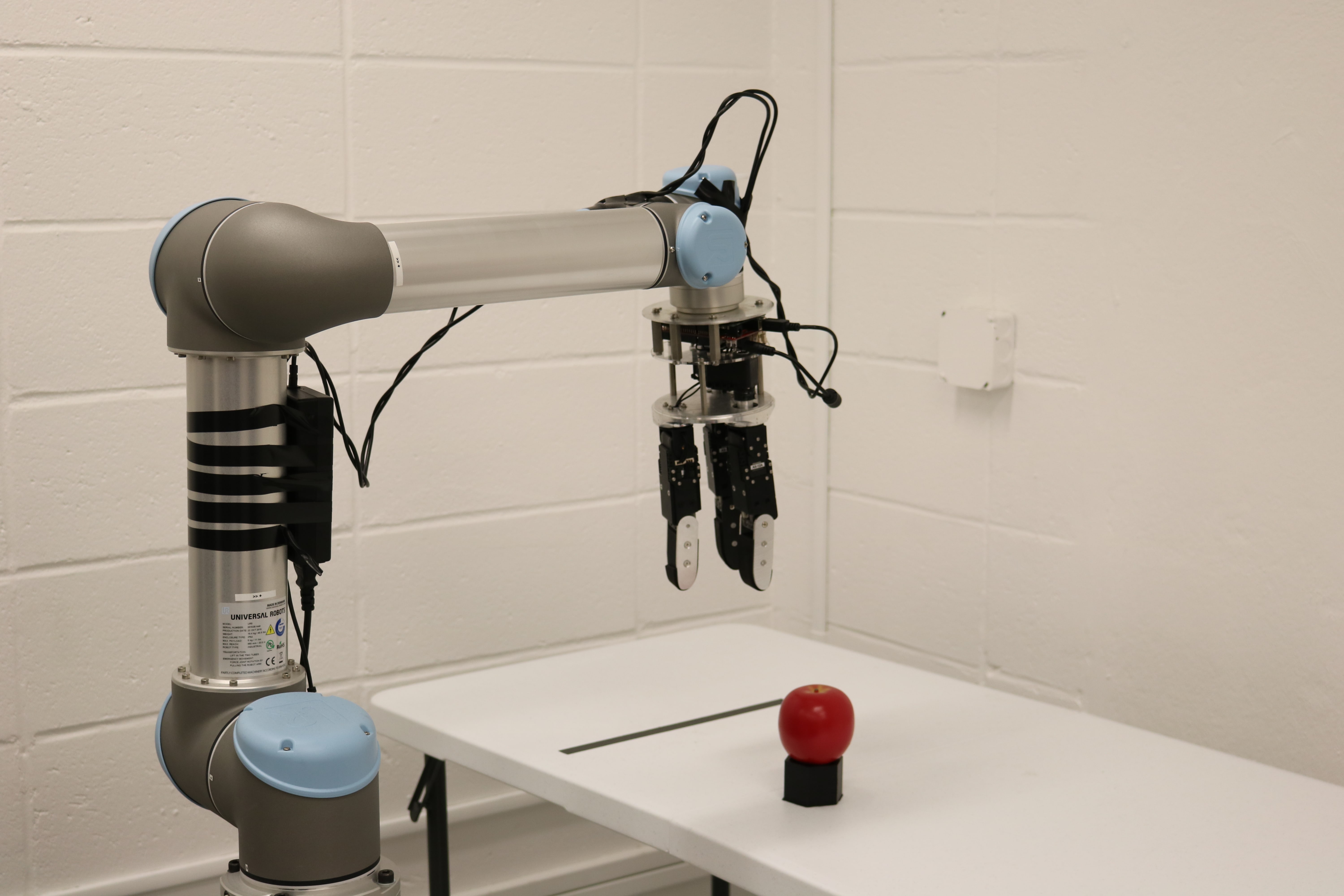}
\includegraphics[width=.42\linewidth]{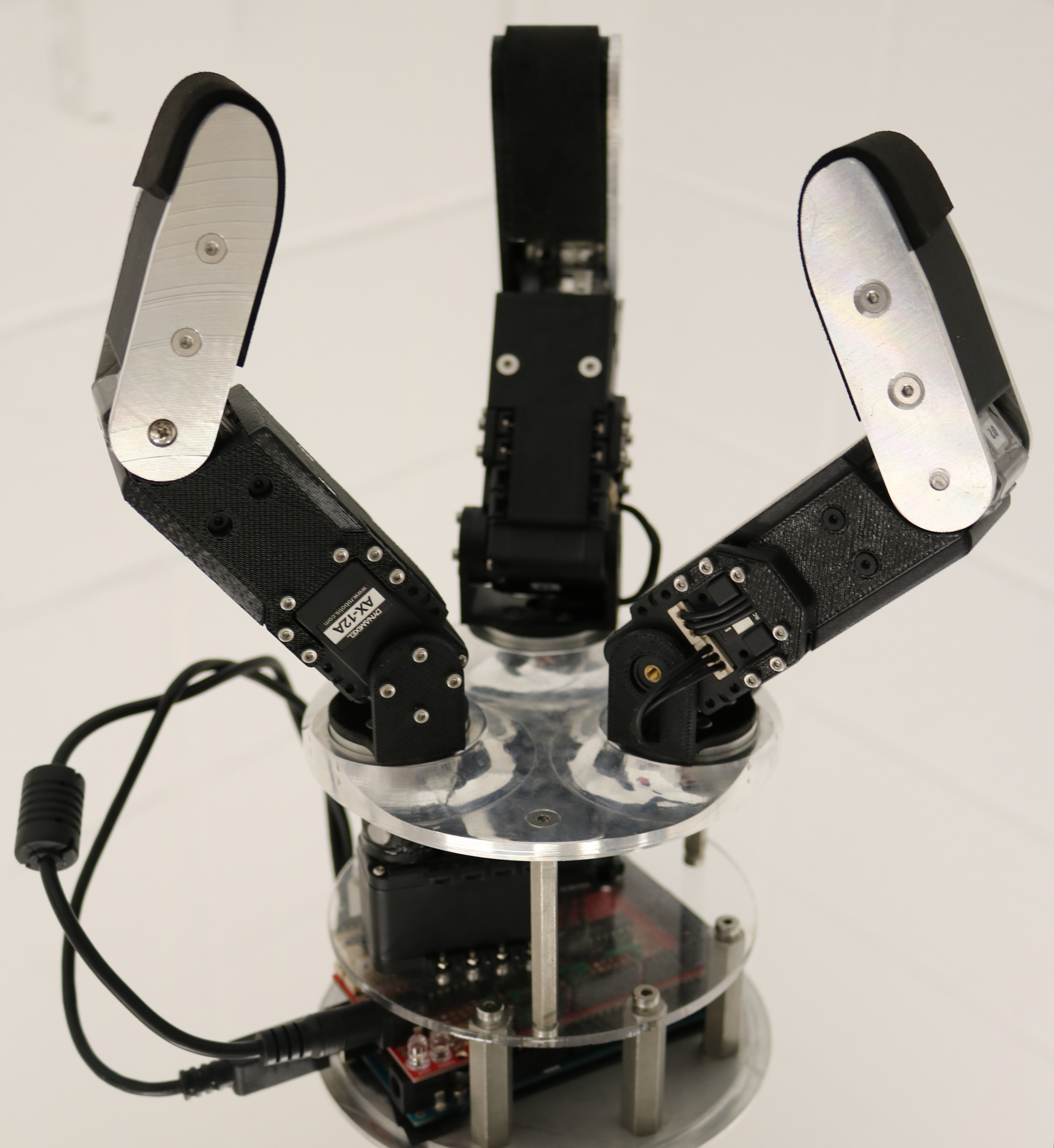}
\end{tabular}
\caption{Left: experimental set-up with UR5 arm, robotic hand and table. Right: close-up of the custom robotic hand used in experiments.}
\label{setup}
\vspace{-4mm}
\end{figure}

The slave hand of our teleoperation system is a custom-built
robotic hand with kinematics similar to the Schunk SDH. It
has a thumb finger and two opposing fingers; all three fingers have two
links. The hand is fully actuated and the two opposing fingers adduct
independently.

We attached the slave hand to a Universal Robot (UR5) arm. The
UR5 stands in front of a table where the grasping objects are
placed one at a time during testing. Figure~\ref{setup} shows
our setup, and a close-up of the slave hand.

The novice teleoperator stands next to the robot and teleoperates
based on visual feedback. Attached to the user's hand is an Ascension 3D Guidance
trakSTAR\texttrademark system,
which tracks hand position and orientation. The UR5 follows the
position and the orientation of the trakSTAR with a cartesian
controller. We control the robotic hand using either joint,
fingertip or teleoperation subspace mapping.

Subjects participated in two testing sessions. During the first
session, we presented the subjects with the subspace and joint
mapping teleoperation methods. During a second session, we presented subjects 
with the fingertip mapping control. The order in which the joint mapping and 
subspace mappings were presented to the users was randomized; however, all subjects
performed the fingertip mapping last. Nominally, subjects should have been able 
to use the fingertip mapping faster because they were already
familiar with the arm and the hand by the second session.

During the first session, the teleoperator wore a data glove (a Cyberglove) 
with a trakSTAR sensor attached to the back of the hand. The Cyberglove provided 
the joint angles of the human hand to the mapping control method. The trakSTAR 
provided hand position and orientation to the UR5 controller.

During the second session, the teleoperator wore a trakSTAR sensor attached to 
the back of the hand, as well as additional sensors attached to three of the 
fingers. All four sensors were used for the fingertip mapping and the sensor 
on the back of the hand again provided hand position and orientation to the 
UR5 controller.

\subsection{Testing}
\begin{figure}[t]
\centering
\begin{tabular}{r}
\includegraphics[width=.75\linewidth]{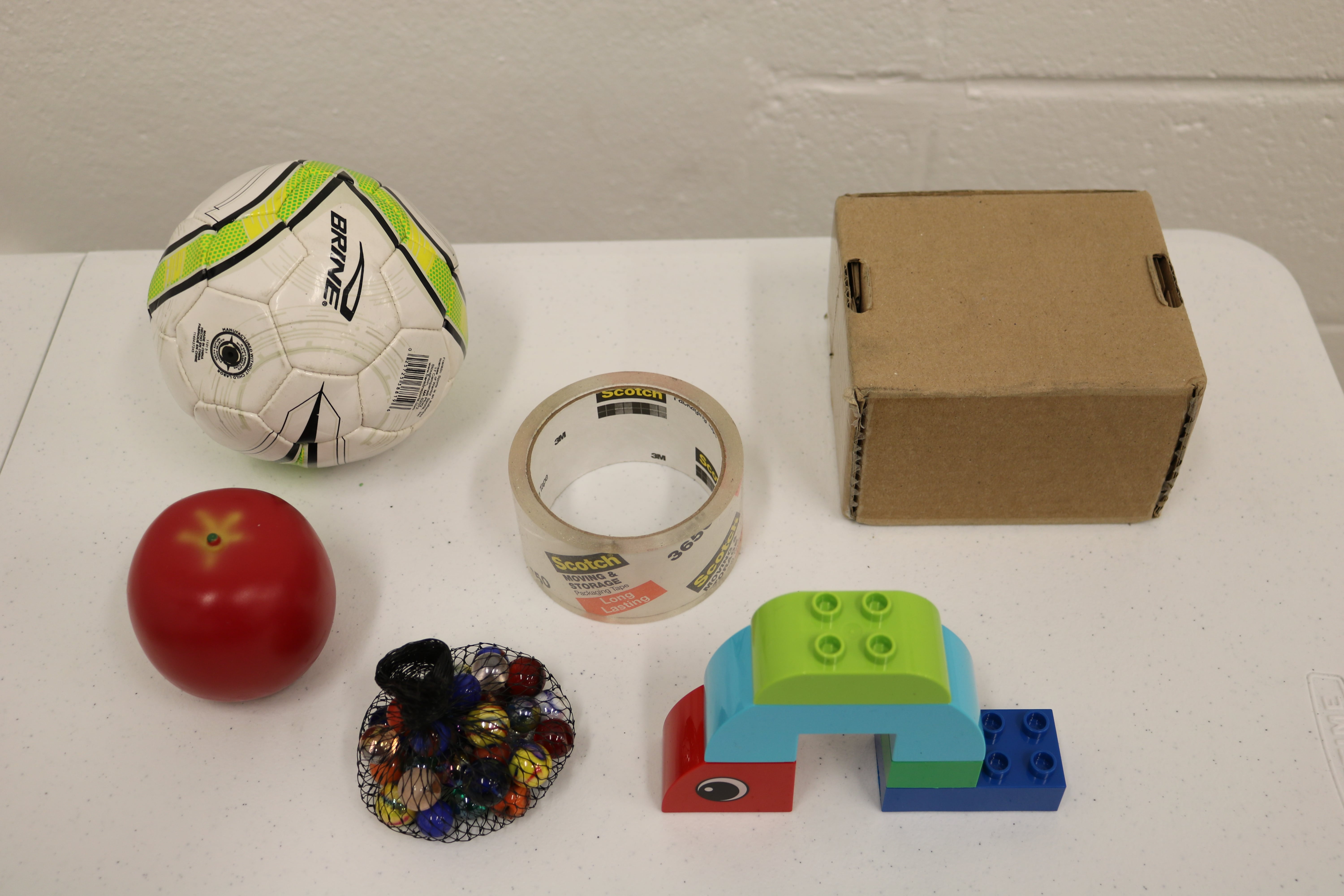}
\end{tabular}
\caption{Objects used in pick and place tasks}
\label{object_set}
\vspace{-4mm}
\end{figure}

We placed a series of objects on the table one at a time and asked 
subjects to pick them up and move them across a line 0.3 meters away.
Shorter objects were placed on a stand so as to facilitate easier grasping with the
large robotic fingers. The objects for the pick and place tasks
were: a box, a ball, a stack of Legos, a roll of tape, a plastic
apple and a mesh bag of marbles (Figure~\ref{object_set}). We selected these object to
illustrate a variety of grasping types. We did not instruct the users 
how they should grasp the objects. After the subject placed
an object in the designated area, we reset the UR5 to a neutral
position before the next pick and place task.

At the beginning of each session, the subject was given three minutes
to move the arm, but not the hand. We did not want the user's unfamiliarity 
with the arm to bias the results towards the second control method during the 
first session. Furthermore, the first object the user attempted to grasp (the 
box) was labeled as a training object and not 
included in the evaluation.

We explained the mapping method to the teleoperator immediately before they 
were asked to pick and place objects using that control.

\subsection{Baseline Comparison: Fingertip Mapping}

We chose fingertip mapping as a comparison because it
is applicable to precision grasps, particularly with smaller objects. 
The fingertip mapping method was designed as follows: first, we
found the cartesian positions of the thumb, index, and ring fingers
of the human hand with respect to the wrist by attaching trakSTAR sensors 
to each of the listed fingers and to the back of the wrist. We calculated 
transforms between the wrist and finger sensors to find the finger
position in the hand frame. We multiplied these positions by a
scaling factor of 1.5, the ratio between an average human finger and one 
of the robotic fingers. We rotated the positions
from the human hand frame into the robotic hand
frame. We translated the coordinates from the robotic hand frame 
into the finger frame to find the desired robotic fingertip positions.
Finally inverse kinematics determined the joint angles which placed the
fingertips at these positions. This process is documented elsewhere~\cite{geng2011}.

\subsection{Baseline Comparison: Joint Mapping}

\begin{table}[]
\centering
\caption{Mapping from the Cyberglove to the Custom Robotic Hand}
\label{joint_mapping_table}
\begin{tabular}{  C{.075\linewidth}  C{.325\linewidth} | C{.075\linewidth}  C{.34\linewidth} }
  \multicolumn{2}{c|}{\textbf{Cyberglove Sensor}}      & \multicolumn{2}{c}{\textbf{Robotic Hand Joints}}         \\   
  Joint Label & Name & Joint Label & Name \\ \hline
  a & 	Thumb adduction 			& 1 & 	Thumb proximal flexion \\ 
  b &	Thumb distal flexion 	& 2 & 	Thumb distal flexion \\ 
  e &	Index/Middle adduction 	& 3 & 	Finger 1 adduction \\   
  c &	Index proximal flexion 	& 4 & 	Finger 1 proximal flexion\\ 
  d &	Index medial flexion 	& 5 & 	Finger 1 distal flexion\\ 
  e &	Index/Middle adduction 	& 6 & 	Finger 2 adduction \\ 
  f & 	Middle proximal flexion 	& 7 & 	Finger 2 proximal flexion \\ 
  g & 	Middle medial flexion 	& 8 & 	Finger 2 distal flexion \\ 
\end{tabular}
\vspace{-2mm}
\end{table}

\begin{figure}[t]
\centering
\begin{tabular}{r}
\includegraphics[trim=5cm 3cm 5cm 2.5cm, clip,width=1\linewidth]{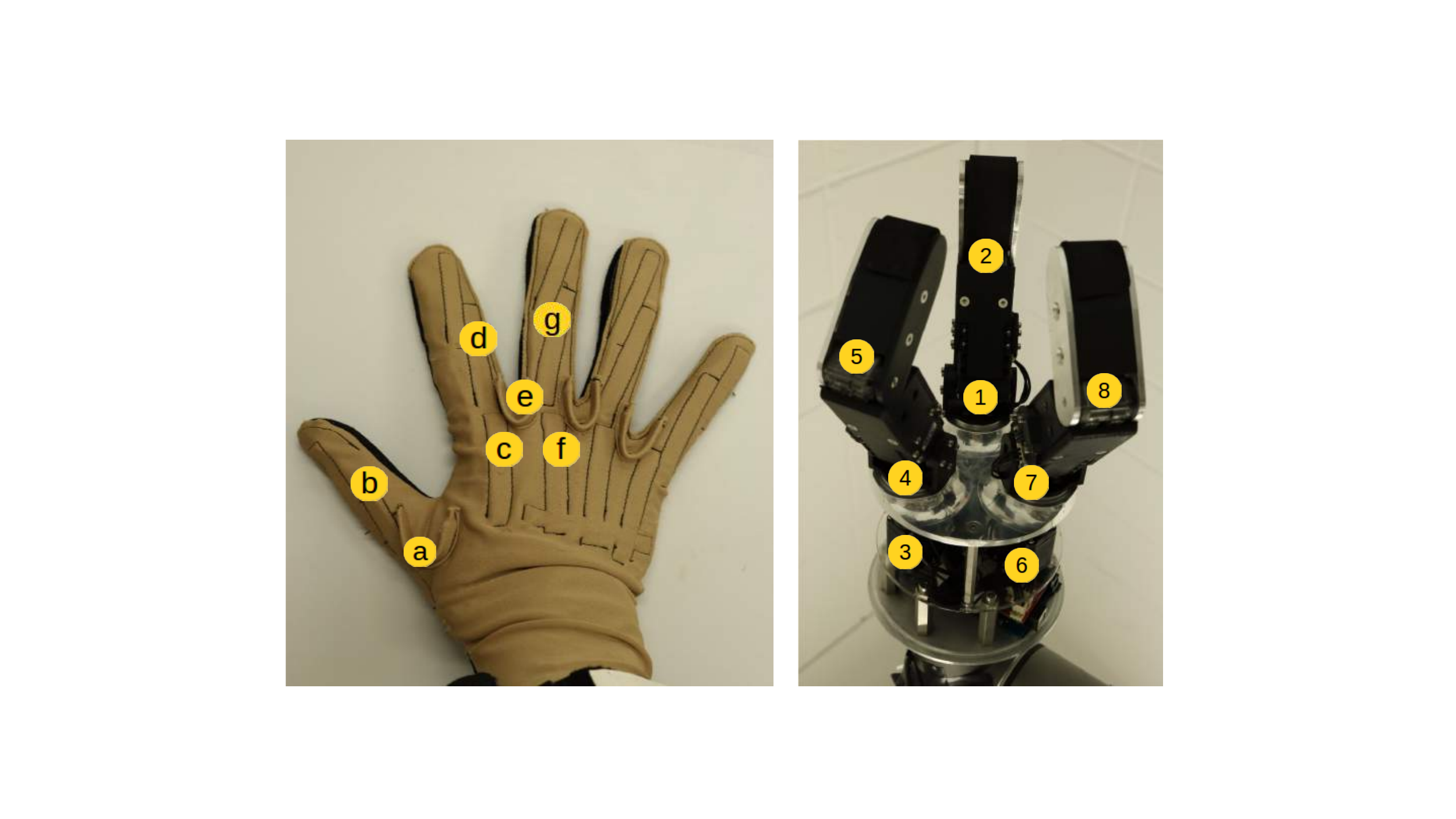}
\end{tabular}
\caption{Joint labels for joint mapping}
\label{joint_mapping_illustration}
\vspace{-4mm}
\end{figure}

We chose joint mapping as the second comparison method
because it is applicable to power grasps. We also 
predicted that explicit control over individual joints of the robotic 
fingers would be intuitive for novice users.
Rosell, et al. teleoperated the Schunk SDH using joint mapping by 
assigning the joints of all fingers of the robot hand to correspond the 
joints of one finger of the human hand, with finger adduction set to a
constant value~\cite{rosell2014}. We instead chose to map each joint of
the robot finger to a separate joint of the human hand so as not to limit
the subject's ability to perform different grasp types.

To implement joint mapping, we assigned each of the
robot joints to a corresponding human hand joint. This mapping can be found in 
Table~\ref{joint_mapping_table} and Figure~\ref{joint_mapping_illustration}. 
We rotated the joint angles of the human hand received from the 
Cyberglove so they aligned with the 
robotic hand, and set the joints of the robot hand to these rotated angles.

Preliminary tests showed teleoperation is difficult if the
robot thumb's proximal joint maps to the human thumb's metacarpophalangeal (MCP) 
joint. We therefore mapped the robot thumb's proximal joint to the human thumb's 
adductor.

\begin{table}[t!]
\caption{Average time (in seconds) to pick and place completion}
\label{experimental_results_table}
\vspace{-3mm}
\centering
\begin{tabular}{C{.15\linewidth} |  C{.05\linewidth}  C{.05\linewidth}   C{.05\linewidth}   C{.05\linewidth}   C{.1\linewidth} | C{.08\linewidth}}
							& \multicolumn{5}{c}{Object}       \\
			Mapping Method	& Ball			& Legos			& Apple			& Tape			& Marbles*		& Average	\\ \hline
			Fingertip 		& 64.67			& 31.83			& 37.05			& 47.02			& 153.6			& 62.27		\\	
 			Joint 			& 23.09			& 45.43			& 51.53			& 27.25			& 157.21			& 56.66		\\
 			Subspaces 		& \textbf{19.22}	& \textbf{22.27}	& \textbf{23.97}	& \textbf{22.63}	& \textbf{33.4}	&\textbf{27.52}	\\
\end{tabular}
\begin{tabular}{l}
\vspace{-3mm}
\\
* denotes that the average was calculated with a smaller sample size \\
for some of the mapping methods of that object.
\end{tabular}
\vspace{-3mm}
\end{table}
\begin{figure}[t!]
\centering
\begin{tabular}{r}
\includegraphics[trim=4cm 15.5cm 3cm 4.4cm, clip,width=1\linewidth]{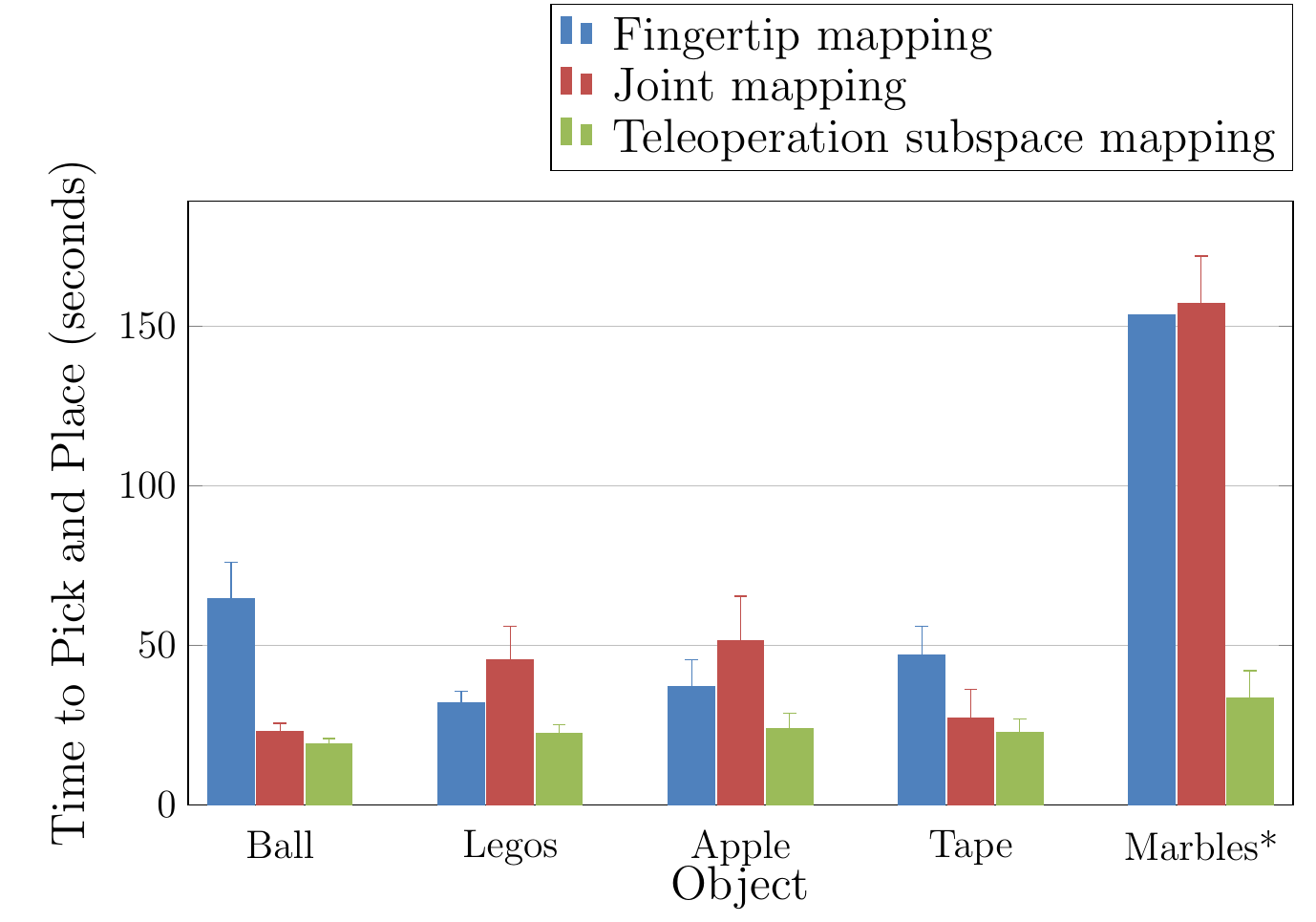}
\vspace{-3mm}
\end{tabular}
\caption{Average time (in seconds) to pick and place objects using different teleoperation controls. * denotes that the average was calculated with a smaller sample size for some of the mapping methods of that object.}
\label{experimental_results}
\vspace{-4mm}
\end{figure}

\subsection{Results}

We timed how long it took for the user to pick and place each object. If the 
user did not complete the task in four minutes, they were considered to be 
unable to pick up the object.

All five novices picked up the ball, legos, apple, and tape in the allotted 
time. All of the users picked up the marbles using subspace mapping. However, 
in the alloted time, only two users picked up the marbles using joint mapping 
and only one user picked up the marbles using fingertip mapping. We therefore 
calculated the average time to pick up the marbles from a smaller sample size 
for the fingertip and joint mappings (denoted by an asterisk in the results figures).

Table~\ref{experimental_results_table} and Figure~\ref{experimental_results} 
show the average time a novice took to pick and place an object using each of 
the control methods. We show our results numerically and graphically. 

We found that novices using fingertip mapping took 2.75 times longer to pick 
and place an object than they took using teleoperation subspace mapping. On 
average, novice teleoperators using joint mapping took 2.51 times longer to 
complete a task than they took using subspace mapping.

For larger objects, like the ball and the tape, fingertip mapping took on 
average 2.75 times longer than subspace mapping, whereas joint mapping only 
took 1.2 times longer. For smaller objects, like the legos and the apple, 
fingertip mapping took 1.49 times longer than subspace mapping, whereas joint 
mapping took on average 2.09 times longer than subspace mapping.

\section{Discussion}
The results show that teleoperation is possible using $T$ as an 
intermediary between the pose spaces of two dissimilar hands. This result 
has several meanings: first, it proves that our projection algorithm and 
our methods for calculating the projection, while simple, are sufficient to 
enable teleoperation. Second, it proves that $T$ is relevant to 
teleoperation for at least the human hand and the custom robotic hand used 
in our experiments. For these two hands, the teleoperation subspace 
encapsulates the range of motion needed to teleoperate the slave hand with 
the human hand.

Our experiments also show that teleoperation subspace mapping allows novice 
users to complete tasks more quickly than they are able to using either 
joint mapping or fingertip mapping. A novice user asked to pick up an 
object will, on average, complete the task 2.75 times slower using 
fingertip mapping and 2.51 times slower using joint mapping than they would 
using teleoperation subspace mapping. Using time to task completion as a 
metric for intuitiveness, these experiments prove that our subspace mapping 
is more intuitive for novice users than state of the art teleoperation 
mappings.

Fingertip mapping is applicable to precision grasps and joint mapping is 
applicable to power grasps~\cite{chattaraj2014}. Our experiments confirm this 
holds true for novice users. For larger objects, like the ball and the 
tape, which were selected to illustrate power grasps, joint mapping allowed 
the user to complete the task in half the time of fingertip mapping. For 
smaller objects, like the stack of Legos and the apple, which were selected 
to demonstrate precision grasps, the reverse was true - joint mapping took 
1.43 times longer than fingertip mapping. Subspace mapping outperformed 
fingertip mapping and joint mapping for both precision grasps and 
power grasps, which shows that it is versatile 
enough to be applicable to different grasp types.

Overall, our experiments showed teleoperation subspace mapping to be faster 
and more versatile when presented with a variety of objects than state of 
the art mapping methods. 

Our method makes several assumptions in order to teleoperate. Guided by 
postural synergies, we assume that the teleoperation subspace we have 
defined encapsulates all of the information needed to teleoperate a slave 
hand with only three basis vectors, and that mapping between joint space 
and teleoperation subspace is linear. We also assume that the user will be 
able to use intuition to generate the three variables needed for each hand 
to create the mapping from joint space to teleoperation subspace. In this 
work, we have provided a clear guideline so the user will be able to 
calculate these variables, and the process we describe is fairly simple. If 
the hand is so non-anthropomorphic that even human intuition cannot find a 
clear mapping into teleoperation subspace, our method will not apply.

Despite these assumptions, the potential applications of our method are 
broad. Unlike other teleoperation methods which use low dimensional 
subspaces, our method does not require the user to generate a large number 
of corresponding poses for the human and the robot. For each hand, the user 
must only provide three variables to enable teleoperation. In our method, 
the mapping between $T$ and pose space is independent of the master-slave 
pairing. If the mapping variables for multiple robots have been defined, 
then each human teleoperator only needs to provide the mapping for their 
specific hand (through a set of calibration poses) in order to teleoperate 
any of the robotic hands. It also is worth noting that, although we 
describe teleoperation in the context of a human controlling a robot, there 
is nothing in our method which requires this, and our method could 
theoretically be used to map poses between two robotic hands.

\section{Conclusions and Future Work}
In this paper, we propose an intuitive, low dimensional mapping between the 
pose spaces of the human hand and a non-anthropomorphic robot hand. We 
present teleoperation subspace as an intermediary between pose spaces of 
different hands. Projecting from pose space of the master hand into 
teleoperation subspace, and then projecting from teleoperation subspace 
into the pose space of the slave hand will enable teleoperation where the 
two hands make similar poses around a scaled object. 

Our experiments show that the proposed teleoperation subspace is indeed 
relevant to teleoperation for at least two hands and that it can enable 
real-time teleoperation of a non-anthropomorphic hand. We also show that 
teleoperation subspace allows novice teleoperators to pick and place 
objects faster than state of the art teleoperation methods.

In the future, we would like to:
\begin{itemize}
\item Test our method with different robotic hands to see if teleoperation 
subspace is relevant to teleoperation for other hands with different 
kinematics.
\item Automate building the projection between pose space and teleoperation 
subspace. Even though we present a simple method to calculate this 
projection, automating the process would make our method more accessible.
\item Explore if the low dimensionality of teleoperation subspace can be 
leveraged to allow for teleoperation using different control inputs, like, 
for example, EMG signals.
\item Study the sensitivity of the proposed mapping to changing parameters, 
such as different origin poses and joint maxima and minima.
\end{itemize}

\bibliographystyle{IEEEtran}
\bibliography{bib/teleoperation}

\end{document}